# Scheduling Distributed Flexible Assembly Lines using Safe Reinforcement Learning with Soft Shielding

Lele Li[1] , Liyong Lin[1*]
1. Contemporary Amperex Technology Co., Limited, Ningde, Fujian Province, China
LiLL03@catl.com, LinLY02@catl.com
Corresponding Author: Liyong Lin Email: LinLY02@catl.com

*Abstract*—**Highly automated assembly lines enable significant productivity gains in the manufacturing industry, particularly in mass production condition. Nonetheless, challenges persist in job scheduling for make-to-job and mass customization, necessitating further investigation to improve efficiency, reduce tardiness, promote safety and reliability. In this contribution, an advantage actor-critic based reinforcement learning method is proposed to address scheduling problems of distributed flexible assembly lines in a real-time manner. To enhance the performance, a more condensed environment representation approach is proposed, which is designed to work with the masks made by priority dispatching rules to generate fixed and advantageous action space. Moreover, a Monte-Carlo tree search based soft shielding component is developed to help address long-sequence dependent unsafe behaviors and monitor the risk of overdue scheduling. Finally, the proposed algorithm and its soft shielding component are validated in performance evaluation.**

## I. Introduction

In today's manufacturing industry, digitization, automation and decision intelligence play significant roles in promoting production efficiency [1]. The flexible assembly lines have been extensively adopted in make-to-job (MTO) as one of the most efficient production forms. Usually, MTO indicates multi-variety, lot dispatching, frequent changeovers, and requirement of flexibility of production [2]. While the flexibility of assembly line enables a great diversity of products, it also poses new challenges in job scheduling, parts feeding, load balancing, and inventory optimizing [3, 4].

Over the last decade, decision intelligence on flexible assembly line scheduling has been widely explored, and innumerable research works have been carried out in this field. J. Mumtaz et al. [5] investigated a multi-level planning and scheduling problem of parallel printed circuit board assembly lines. With the goal of maximizing net profit, a hybrid spider monkey optimization approach was developed to handle this problem. To address the integrated optimization problem of aircraft assembly scheduling and its preventive maintenance, Q. Yan and H. Wang [6] proposed a mixed integer linear programming (MILP) model and then a double-layer Q-learning based reinforcement learning approach to carry out joint decision-making. Considering energy-saving control and various disruptions, Y. Li et al. [7] built an analytical model to systematically quantify production loss, and with that a Markov decision process (MDP) based energy-saving control method was developed to reduce energy consumption while maintaining desirable productivity. T. Yu et al. [8] studied a sequence optimization problem in the human-robot collaborative assembly shop, and a reduced collection of the components used in Alphago Zero was adopted to conduct real-time task planning.

In more complex scenarios, D. Li et al. [9] investigated a balancing optimization problem of mixed-model two-side assembly line, and proposed a novel multi-objective hybrid imperialist competitive algorithm to solve this problem, of which a late acceptance hill-climbing method was encapsulated as the local search operator to carry out variable neighborhood search. To achieve high throughput with minimal idle time in a two-type door manufacturing line, C. Zhao et al. [10] formulated a renewal model to evaluate production performance and quantify the optimal control thresholds, and the sensitivity analysis considering setup time change, machine reliability variation, and demand fluctuation was performed to study the monotonicity of production control parameters. Taking partition delivery and robot energy limitation into account, X. Ma and X. Zhou [11] presented a multi-objective mathematical robot scheduling model in mixed-model assembly lines; furthermore, the authors also developed a new nondominated sorting genetic algorithm II with variable neighborhood search, initialization method, crossover operation, and local search strategies. Z. Bao et al. [12] studied a balancing problem of the aircraft final assembly line and formulated a mixed integer linear programming model. A two-stage heuristic approach with the objectives of balancing cycle time and minimizing total cost is proposed to solve the problem. Considering both processing stage and assembly stage in the mixed-model assembly job shop, L. Cheng et al. [13] developed a mathematical model based on the temporal and spatial links between the two stages. In addition, an adaptive simulated annealing algorithm is employed to optimize lot streaming problem in the processing stage, and a Q-learning based selection operator was designed to improve local search performance. Out of the concern of in-process safety, C. Zhu et al. [14] and R. Zhang et al. [15] formulated the robot motion planning and control problem for the human-robot collaboration assembly respectively, and both developed deep reinforcement learning methods with safety control.

In this paper, the job scheduling problem (inherently involving two subproblems: task allocation and sequence optimization) of distributed flexible assembly lines is studied. The contributions of this paper are stated as follows. Firstly, we formulate the job scheduling problem of flexible assembly lines into an MILP model, and the changeover cost and overdue inventory cost are considered as the objective. For real-time application, an advantage actor-critic based deep reinforcement learning method with newly-developed feature representation is employed to make dynamic dispatching decisions. Moreover, a Monte Carlo tree search method based soft shielding component is proposed to address the unsafe behavior caused by long sequence dependent overdue dispatching. To our best knowledge, few works have been conducted in this field to study the long sequence dependent overdue dispatching problem.

## II. Problem Statement and Formulation

### A. Problem Statement

There is a production entity composed of several assembly lines, and each line is designed to complete different jobs. In our setup, each job consists of only one work order. Each job has variable facility capacity plans (FCPs) in different lines. For each job, it can be allocated in a line only if the line is able to provide the production ability accordingly. In addition, the changeover operations occur when switching between different jobs, and the changeover timespan varies according to the neighboring jobs and the line they are allocated. As each job has its demand time window, it should not get completed in advance or get delayed so as to reduce extra inventory cost.

There can be non-regular job changeover situations, which are detailed as follows. First of all, the changeover timespan may be extremely long. For example, the whole changeover may span several days, and no job can get started during this period. Furthermore, the lot splitting (or lot streaming) is not supported as this strategy would lead to more changeovers. Overdue scheduling can happen as plenty of jobs may occur at the same demand window and only partial flexibility of assembly lines is considered.

In this problem, the decisions on the subproblems of job allocation and sequence optimization are specified concerning the hierarchical objectives of minimizing the total changeover timespan while reducing tardiness.

### B. Mathematical Formulation

According to the above statement, a position-based mixed-integer linear programing model for the problem is formulated as follows, the mathematical notations are tableted in TABLE I.

TABLE I. MATHEMATICAL NOTATION AND DEFINITION.

| Notation | | Definition |
|---|---|---|
| index | $i$ | Index of job, $i \in \{1, \ldots, I\}$ |
| | $j$ | Index of line, $j \in \{1, \ldots, J\}$ |
| | $r$ | Index of task on the line, $r \in \{1, \ldots, R\}$ |
| | $d$ | Index of date, $d \in \{1, \ldots, D\}$ |
| constant | $C_{i,i',j}$ | "Continuous", changeover timespan from $i$ to $i'$ $at$ line $j$, and $C_{0,i,j}$ indicates the first changeover from some job to job $i$ as the first task of line $j$ |
| | $D_i$ | "Integer", the demand lot of job $i$ in the day $U_i$ |
| | $R_{i,j}$ | "Continuous", production quantity per hour for job $i$ in line $j$ |
| | $P_{i,j}$ | "Binary", whether job $i$ could be processed at line $j$ |
| | $U_i$ | "Integer", the day (has 24h) when job $i$ is to be processed |
| | $Setup_j$ | "Continuous", The setup time of line $j$ |
| variable | $s_{r,j}$ | "Continuous", the start time of $r$th task at line $j$ |
| | $\alpha_{i,r,j}$ | "Binary", whether job $i$ is the $r$th task at line $j$ |
| | $t_{r,j}$ | "Continuous", changeover timespan of $r$th task at line $j$ |
| | $o_{i,r,j}$ | 'Continuous', overtime of job $i$ when dispatched to the $r$th task at line $j$, it is 0 if job $i$ is not the $r$th task at line $j$ |

objective expression:

$$\text{minimize: } obj1{:}\ \textstyle\sum_{r,j} t_{r,j};\ obj2{:}\ \sum_{i,r,j} o_{i,r,j}$$

subject to:

$$\textstyle\sum_i \alpha_{i,r,j} \le 1, \forall (r,j) \quad (1)$$

$$\textstyle\sum_{r,j} \alpha_{i,r,j} = 1, \forall i \quad (2)$$

$$\textstyle\sum_r \alpha_{i,r,j} \le P_{i,j}, \forall (i,j) \quad (3)$$

$$s_{r,j} + M(1 - \alpha_{i,r,j}) \ge 24(U_i - 1), \forall (i,r,j) \quad (4)$$

$$s_{1,j} - t_{1,j} \ge Setup_j, \forall j \quad (5)$$

$$s_{r,j} + \frac{D_i \alpha_{i,r,j}}{R_{i,j}} - M(1 - \alpha_{i,r,j}) \le 24U_i + o_{i,r,j}, \forall (i,r,j) \quad (6)$$

$$o_{i,r,j} \ge 0, \forall (i,r,j) \quad (7)$$

$$t_{r,j} + M(2 - \alpha_{i,r,j} - \alpha_{i',r-1,j}) \ge C_{i,i',j}, \forall (i,i',r,j) | \{(i \ne i') \wedge (r \ge 2)\} \quad (8)$$

$$t_{1,j} + M(1 - \alpha_{i,1,j}) \ge C_{0,i,j}, \forall (i,j) \quad (9)$$

$$s_{r,j} + \frac{D_i \alpha_{i,r,j}}{R_{i,j}} - M(1 - \alpha_{i,r,j}) + t_{r+1,j} \le s_{r+1,j}, \forall (i,r,j) | (r \le R - 1) \quad (10)$$

This MILP model is to minimize the total changeover timespan while reducing the total tardiness. Linear expression (1) puts the constraint that the $r$th task of line $j$ can only accept one job at most. Equation (2) indicates job $i$ has to get dispatched to one line (not allowing lot splitting). Equation (3) poses a logistic constraint that job $i$ is possible to be processed at line $j$ only if line $j$ meets the condition (i.e., $P_{i,j} = 1$), otherwise job $i$ cannot be allocated to line $j$ (i.e., $P_{i,j} = 0$). Equation (4) and (5) present a joint constraint on the start time of $r$th task at line $j$, and equation (6) puts a similar constraint upon the end time. Equation (7) gives the low bound of continuous variables $o_{i,r,j}$ . Equation (8) and (9) constrain the changeover timespan of $r$th run at line $j$, and equation (10) gives the sequential constraint on the start time of neighboring tasks.

## III. METHOD

### A. Representation for MDP

According to the problem statement (see section II.A), the distributed flexibly assembly line scheduling problem can be viewed as a flexible job-shop scheduling problem with single operation (FJSP-SO), i.e., a simplified FJSP. From the perspective of solution construction, the whole process consists of a group of partial solutions of dispatching operations step by step in a sequential way. For one operation, there are multiple dispatching options, each substitute option would result in different states at the upcoming step, and the future state depends only on the current state. With this viewpoint, a MDP based deep reinforcement learning (DRL) method is possible to be established for the problem.

- *State Features*

The partial solution grows up dynamically from the scratch when making dynamic dispatching decisions. In this process, the dispatched operations have become part of the whole environment, thus the following dispatching operations can only be generated from the remaining operations, which is called residual scheduling [16]. With this in mind, the state of each line and the dependent information about their potential following options can be given and updated in a Markov decision process.

The real situation of the production system with multiple assembly lines is quite complex in practice, and thus only regular setups as well as the problem-related information is presented in this section. Typically, we need to specify the status, properties, and requirements of each line and operation together with their combined status to detail the whole environment state. When encoding the above information, an obvious problem is that heterogeneous data with multiple channels and variable attention weights may arise, it usually requires sophisticated encoding representation designs. Therefore, a novel condensed representation approach of environment state features is proposed in this paper.

Since the decisions are all made in the temporal dimension, the state representation can be transformed into a more condensed form (illustrated in Fig. 1).

| | | Line 1 | Line 2 | Line 3 | Line 4 | Line 5 | Line 6 | ... |
|---|---|---|---|---|---|---|---|---|
| Occupancy at *t*th step | | 0.122 | 0.000 | 0.000 | 0.079 | 0.032 | 0.158 | ... |
| Mask played by PDRs | SCT increment | 0.000 | 0.029 | 0.000 | 0.000 | 0.000 | 0.000 | ... |
| | SPT increment | 0.147 | 0.000 | 0.000 | 0.000 | 0.000 | 0.000 | ... |
| | LPT increment | 0.000 | 0.000 | 0.350 | 0.000 | 0.000 | 0.000 | ... |
| | EDD increment | 0.000 | 0.029 | 0.000 | 0.000 | 0.000 | 0.000 | ... |
| | ECT increment | 0.000 | 0.000 | 0.000 | 0.450 | 0.000 | 0.000 | ... |
| | others | ... | ... | ... | ... | ... | ... | ... |

Fig. 1. Representation of state features.

For convenience, we only describe a truncated environment state consisting of several assembly lines and remaining operations. In the temporal dimension, the status of each assembly line can be given through the occupancy degree during its scheduling span or period. Following that, we also need to depict the remaining operations from the viewpoint of temporal dimension. The information indicating the next dispatching options is embedded in the truncated environment state. Therefore, only the operation related environment features are described and encoded, i.e., the information of the target dispatching operations such as the changeover timespan, processing timespan, the completion time, and the start time, etc. However, the key problem is how to enumerate candidate dispatching options with high quality to support decision making.

- Actions

The priority dispatching rules (PDRs) are employed here to detail the description of action space. PDRs can provide advantageous partial solutions compared to random dispatching. Moreover, PDRs based dispatching operations can provide relatively fixed action space, which can be made arbitrarily close to optimal choices by increasing the number of rules used. Furthermore, PDRs naturally give the action space a desired filter tool to avoid homogeneous operations, which is a highly practical technique in MTO scenario.

In this paper, only several time-related PDRs are considered, which are detailed as follows:

(1) The shortest changeover timespan (SCT).
(2) The shortest processing timespan (SPT).
(3) The shortest total consuming timespan (STCT).
(4) The shortest weighted total consuming timespan (SWTCT).
(5) The longest processing timespan (LPT).
(6) The longest total consuming timespan (LTCT).
(7) The earliest due date (EDD).
(8) The earliest completion time (ECT).
(9) The earliest start time (EST).
(10) The shortest relaxation timespan (SRT).

These PDRs make up the mask of real action space, which promotes the computational efficiency while retaining sufficient dependent information to support decision making.

- Reward

The objective of the introduced problem is to minimize the total changeover timespan while reducing tardiness. Thus, the reward design in this DRL method is also given in the temporal dimension. Concerning a specific action $a$, assume the operation is dispatched at line $j$ from job $i$ to $i'$, the step reward $r_a$ is formulated as equation (11), of which $h_a$ is an indicator variable that represents a penalty when action $a$ leads to overdue scheduling and 0 otherwise.

$$r_a = \frac{1}{1 + C_{i,i',j}} - h_a \tag{11}$$

## B. A2C Algorithm

As a policy gradient method, advantage actor-critic (A2C) algorithm employs a policy network $\pi(a|s;\boldsymbol{\theta})$ and a Q-value network $v(s;\boldsymbol{w})$ to estimate the optimal policy function and the optimal value function, respectively (see Fig. 2).

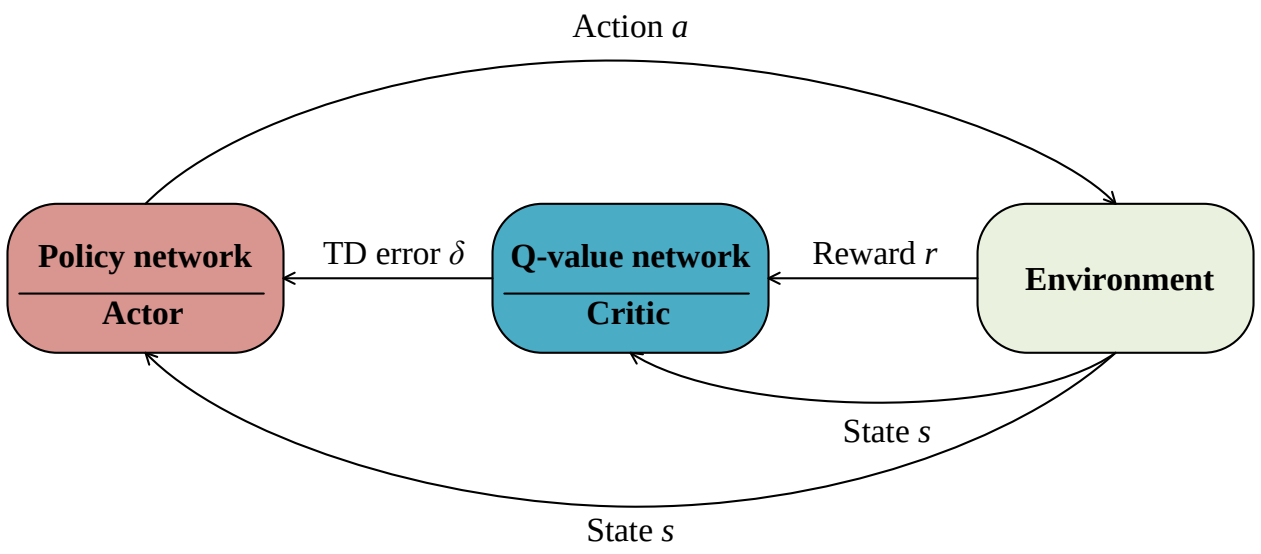

Fig. 2. Relationships of the roles in A2C algorithm.

Consider the whole operation dispatching process, whose MDP trajectory is observed as $S_t$, $A_t$, $S_{t+1}$, $A_{t+1}$, $S_{t+2}$, $A_{t+2}$, … . Let us denote $U_t = \sum_t \eta^t r_t$ as the cumulative future reward at $t$th step, of which $r_t$ is the environment reward at $t$th step, and $\eta$ is a discounted factor. The action-value function (12) and value function (13) are formulated as follows.

$$Q_\pi(s_t, a_t) = E(U_t | S_t = s_t, A_t = a_t) \tag{12}$$
$$V_\pi(s_t) = E_{A_t \sim \pi(\cdot|s_t;\boldsymbol{\theta})}[Q_\pi(s_t, A_t)] \tag{13}$$

With the purpose of increasing $V_\pi(s_t)$ at $t$th step, the objective function is formulated as (14), of which the policy function $\pi(a|s;\boldsymbol{\theta})$ makes an extremely significant role.

$$J(\boldsymbol{\theta}) = E_S[V_\pi(S)] \tag{14}$$

In the training phase, this objective of policy network can be maximized by optimizing the hyper-parameters $\boldsymbol{\theta}$, i.e., $\widehat{\boldsymbol{\theta}} = \max_{\boldsymbol{\theta}} J(\boldsymbol{\theta})$. At each step, parameter $\boldsymbol{\theta}$ can be updated through $\boldsymbol{\theta}_{new} \leftarrow \boldsymbol{\theta}_{old} + \gamma \cdot \nabla_{\boldsymbol{\theta}} J(\boldsymbol{\theta}_{old})$, of which $\gamma$ represents the learning rate, and the policy gradient $\nabla_{\boldsymbol{\theta}} J(\boldsymbol{\theta}) = E_{S\sim p(\cdot|S,A)}\left[E_{A\sim\pi(\cdot|S;\boldsymbol{\theta})}\left[\frac{\partial \ln(\pi(A|S;\boldsymbol{\theta}))}{\partial \boldsymbol{\theta}} \cdot Q_\pi(S,A)\right]\right]$ when the state transition $p(\cdot)$ obeys the steady-state distribution of the Markov chain.

In A2C, it is possible to refine the $\nabla_{\boldsymbol{\theta}} J(\boldsymbol{\theta})$ using policy gradient with baseline. Usually, the value function $V_\pi(S)$ is employed as the baseline. The new policy gradient can be formulated as $\nabla_{\boldsymbol{\theta}} J(\boldsymbol{\theta}) = E_S\left[E_{A\sim\pi(\cdot|S;\boldsymbol{\theta})}\left[\frac{\partial \ln(\pi(A|S;\boldsymbol{\theta}))}{\partial \boldsymbol{\theta}} \cdot (Q_\pi(S,A) - V_\pi(S))\right]\right]$, and $Q_\pi(S,A) - V_\pi(S)$ is the advantage function. According to Bellman Equation, $V_\pi(s_t) = E_{A_t\sim\pi(\cdot|s_t;\boldsymbol{\theta})}\left[E_{S_{t+1}\sim p(\cdot|s_t,A_t)}[R_t + \eta \cdot V_\pi(S_{t+1})]\right]$, of which $R_t$ is the environment reward at $t$th step, and $\eta$ is a discounted factor.

The imitator of $V_\pi(S)$ is a network $v(S;\boldsymbol{w})$, with which we can also perform precise estimation for the cumulative future reward as it considers partial real data $r_t$, i.e., $\hat{y}_t = r_t + \eta \cdot v(s_{t+1};\boldsymbol{w})$, and $\hat{y}_t$ is the temporal difference (TD) target. The loss function of $v(S;\boldsymbol{w})$ is defined with TD error: $L(\boldsymbol{w}) = [\delta_t]^2/2$, and TD error $\delta_t = v(s_t;\boldsymbol{w}) - \hat{y}_t$, then the value network can be updated using $\boldsymbol{w}_{new} \leftarrow \boldsymbol{w}_{old} - \gamma \cdot \nabla_{\boldsymbol{w}} L(\boldsymbol{w}_{old})$. To alleviate the deviation caused by bootstrapping, a newly designed training process for A2C is detailed as follows.

Online A2C algorithm with target network

---

initialize policy network $\pi(a|s;\boldsymbol{\theta}_0)$, value network $v(s;\boldsymbol{w}_0)$, and target network $v'(s;\boldsymbol{w}'_0)$ with normalization
make initial environment state
for $episode = (1, \mathrm{n})$ do
  initialize current step: $t = 1$
  while $t \le finalStep$ do
    take action $a_t \sim \pi(\cdot|s_t;\boldsymbol{\theta})$ on the state $s_t$, get $(s_t, a_t, s_{t+1}, r_t)$
    score state $s_t$ with value network $\hat{v}_t = v(s_t;\boldsymbol{w}_{now})$
    score state $s_{t+1}$ with target network $\hat{v}'_{t+1} = v'(s_{t+1};\boldsymbol{w}'_{now})$
    calculate TD target $\hat{y}'_t = r_t + \eta \cdot \hat{v}'_{t+1}$ and TD error $\delta_t = \hat{v}_t - \hat{y}'_t$
    update value network $\boldsymbol{w}_{new} \leftarrow \boldsymbol{w}_{now} - \gamma_1 \cdot \delta_t \cdot \nabla_{\boldsymbol{w}} v(s_t;\boldsymbol{w}_{now})$
    update policy network $\boldsymbol{\theta}_{new} \leftarrow \boldsymbol{\theta}_{now} - \gamma_2 \delta_t \nabla_{\boldsymbol{\theta}} \ln(\pi(a_t|s_t;\boldsymbol{\theta}_{now}))$
    update target network $\boldsymbol{w}'_{new} \leftarrow \lambda \cdot \boldsymbol{w}_{new} + (1-\lambda) \cdot \boldsymbol{w}'_{now}$
    update current step: $t$ += 1
  end while
  reset environment state
end for

---

## C. Soft Safety Component

In practice, the safety problem (overdue dispatching) may occur quite frequently, for which a soft safe shielding component is proposed. Unusually, it is the long-sequence dependent risk rather than the direct unsafe operation with obvious infeasibility that is considered in this problem, the overdue dispatching and its risk dependencies are shown in Fig. 3.

To address the long-sequence dependent safety problem, a soft shielding component using Monte Carlo method is developed to monitor risk before real decision-making.

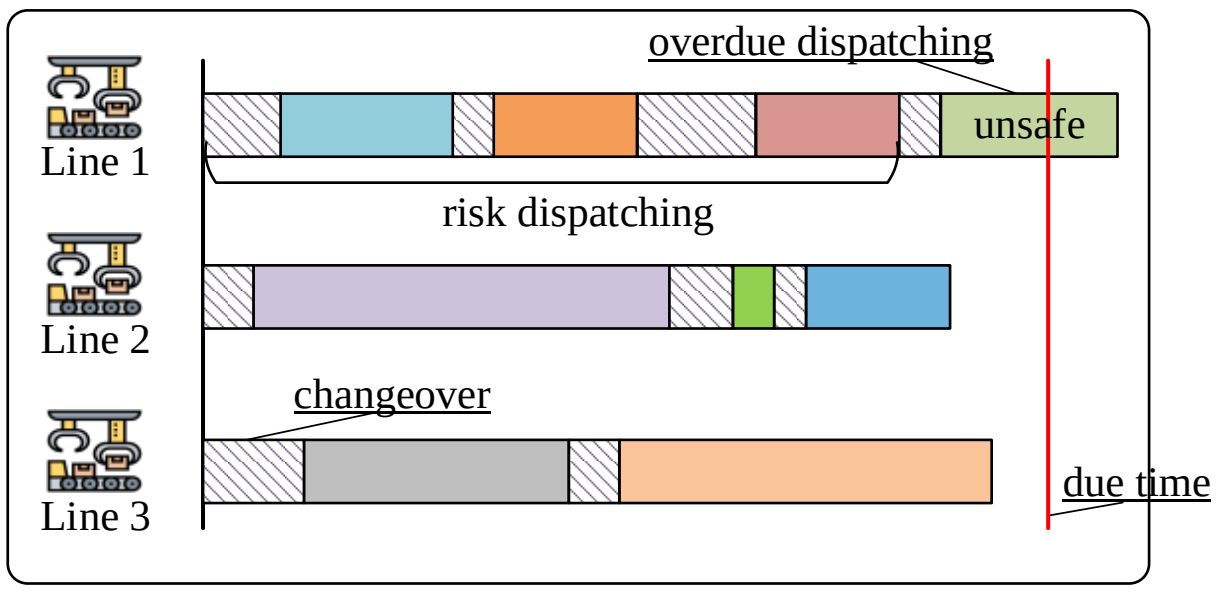

Fig. 3. sequence-depended overdue dispatching.

At the environment state $s_t$, the potential actions as well as their probability values can be enumerated through a well-trained policy network $\pi(\cdot\,|s_t;\boldsymbol{\theta})$. Before actually acting, it is possible to conduct simulated dispatching operations to the end of this episode. In each dispatching simulation, we sample an action from $\pi(\cdot\,|s_t;\boldsymbol{\theta})$ at $t$th step and then virtually carry out the action and evaluate the new state $s_{t+1}$ with $v(s_{t+1};\boldsymbol{w})$.

After state $s_{t+1}$ is reached, we can make continuous simulated dispatching decisions with policy network until there is no task left. This is a stochastic process as each decision-making action is determined through a probability density function $\pi(\cdot\,|s;\boldsymbol{\theta})$. The $k$th simulation can offer a detailed result $u_k$ indicating whether there is any overdue scheduling. We score the action $a_t$ using equation (15).

$$G(a_t) = \frac{\alpha}{n(a_t)+1}\sum_{k=1}^{n(a_t)} u_k + (1-\alpha)\cdot v(s_{t+1};\boldsymbol{w}) + \frac{\beta\cdot\pi(a_t|s_t;\boldsymbol{\theta})}{1+n(a_t)} \qquad (15)$$

Monte Carlo tree search based soft shielding initialize action set: $set_{\pi,s_t}$ and the simulation tools: $\pi(\cdot\,|s;\boldsymbol{\theta})$, $v(s;\boldsymbol{w})$
start $K$ simulations with current state $s_t$:
for $k = (1, K)$ do
  initialize $u_k$
  sample an action $a_t \sim \pi(\cdot\,|s_t;\boldsymbol{\theta})$
  update action sampling frequency: $n(a_t)$ += 1
  virtually carry out action $a_t$, observe a new state $s_{t+1}$
  evaluate the new state $s_{t+1}$: $v(s_{t+1};\boldsymbol{w})$
  initialize current step: $r = k + 1$
  while $r \le finalStep$ do
    sample an action $a_r \sim \pi(\cdot\,|s_r;\boldsymbol{\theta})$ on the observed state $s_r$
      if action $a_r$ results in overdue dispatching do
        $u_k = -1$
        break while
      if $r$ equals $finalStep$ do
        $u_k = 1$
      end if
      update current step: $r$ += 1
  end while
  reset simulation state to $s_t$
end for
end simulation
for $a_t$ in set $set_{\pi,s_t}$:
  calculate score $G(a_t)$ through equation (15)
end for

After simulation, we traverse each action $a_t$ and calculate its score $G(a_t)$ through equation (15), where $n(a_t)$ refers to sampling frequency of action $a_t$ in total $K$ dispatching simulations. $\alpha$ and $\beta$ are hyperparameters determining factor weights. The real dispatching decision is made with $\underset{a_t}{\operatorname{argmax}}\, G(a_t)$.

## IV. PERFORMANCE EVALUATION

### A. Experiment Setting and metrics

In this section, the performance of the proposed DRL scheduling model and the soft safety component is evaluated. Prior to this, a set of recommended model settings and hyperparameters based on experience is provided (see TABLE II).

Since the problem mainly focuses on minimizing the changeover timespan and reducing total tardiness, the related metrics on changeover and overdue scheduling are proposed in this paper; and they are expected to offer an impartial numeric scale for performance evaluation in the next experiments.

Let $C_k$ and $T_k$ denote the changeover and tardiness timespan of job $k$ respectively, and there are totally $K(Inst.p)$ jobs to be dispatched to $n$ lines and $d$ days in the instance $Inst.p$. The daily changeover load (DCL) as well as the daily tardiness load (DTL) can be formulated as equation (16) and equation (17), respectively.

$$DCL|_{Inst.p} = \frac{1}{n\cdot d}\sum_{k}^{K(Inst.p)} C_k \qquad (16)$$

$$DTL|_{Inst.p} = \frac{1}{n\cdot d}\sum_{k}^{K(Inst.p)} T_k \qquad (17)$$

TABLE II. EXPERIENCE VALUES OF MODEL SETTINGS AND HYPERPARAMETERS.

| **Setting item** | **Experience value** |
|---|---|
| network initialization | 1-D Gaussian Function with ($\mu = 0, \sigma = 0.1$) |
| optimizer | Adam |
| learning rate | 3.8E-4 |
| discounted factor $\eta$ | 0.95 |
| weight factor $\lambda$ | 0.7 |
| penalty indicator $h_a$ | {0.28, 0} |
| weight factor $\alpha$ | 0.95 |
| weight factor $\beta$ | 0.33 |
| simulation times | 1.2E3 |

Also, the benchmark instances with middle and large size are prepared, and the problem characteristics are summarized in TABLE III.

TABLE III. BENCHMARK INSTANCES AND THEIR CHARACTERISTICS.

| Instance | Lot Size | Lines | Plan period (day) | Line flexibility | Problem size |
|---|---|---|---|---|---|
| Inst.01 | 61 | 4 | 7 | 1-10 | middle |
| Inst.02 | 88 | 4 | 10 | 1-12 | middle |
| Inst.03 | 121 | 6 | 10 | 1-15 | large |
| Inst.04 | 151 | 6 | 12 | 1-15 | large |
| Inst.05 | 170 | 6 | 14 | 1-15 | large |

### B. *Results and Discussion*

For comparison, all tests were carried out on a laptop with Intel Core i7-1165G7 and 16GB RAM using the benchmark instances. In the experiment, the proposed A2C scheduling model together with its soft safety component were firstly verified, the comparison between the proposed A2C scheduling model and several PDRs was illustrated in Fig. 4.

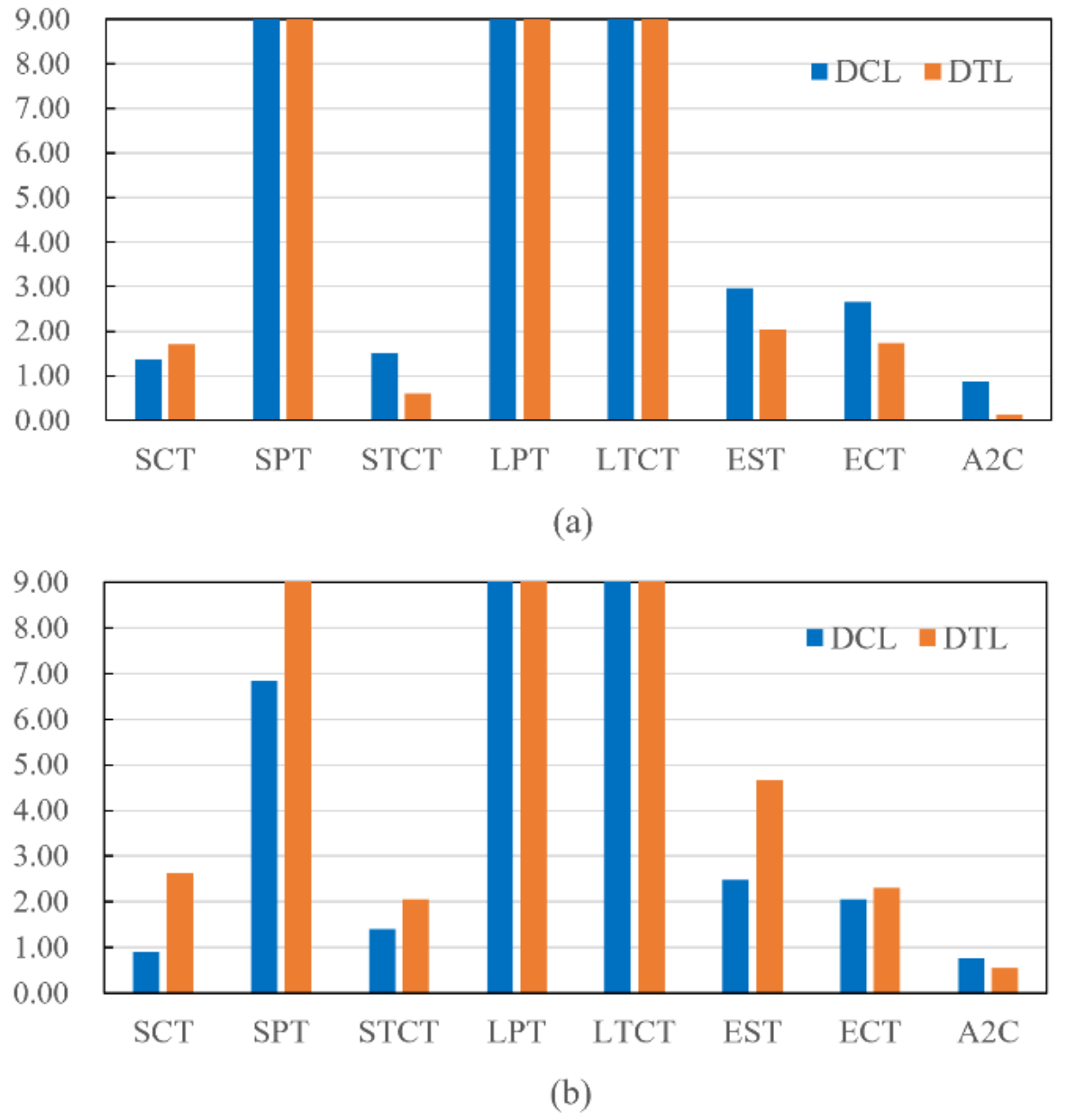

Fig. 4. Comparison between A2C scheduling model and several PDRs on (a) Inst.01 and (b) Inst.03.

To make further verification, the improved niche genetic algorithm (INGA) by [17] and the IBM CPLEX solver (version 22.1.1) running the MILP model (see section II.B) were tested in the same test environment. In more details, the A2C scheduling model was trained until convergence before making dynamic dispatching decisions, and INGA was also run to convergence to provide its final solutions. However, for CPLEX solver, it was allowed to run at most 1 hour for each instance as the solver running a deterministic branch-and-cut method may be extremely time-consuming on the instances with middle to large problem sizes. The optimal outcomes from 5 runs on benchmark instances were documented for each scheduling model, and the corresponding numerical data has been presented in TABLE IV.

TABLE IV. SOLUTION RESULTS OF SEVERAL TESTED ALGORITHMS.

| Inst. | A2C | | A2C with shielding | | INGA | | CPLEX solver | |
|---|---|---|---|---|---|---|---|---|
| | DCL | DTL | DCL | DTL | DCL | DTL | DCL | DTL |
| Inst.01 | 0.868 | 0.124 | 0.939 | 0.100 | 1.080 | 0.091 | 1.210 | 0.163 |
| Inst.02 | 1.107 | 0.982 | 1.357 | 0.265 | 1.206 | 0.312 | 1.781 | 1.026 |
| Inst.03 | 0.756 | 0.539 | 0.905 | 0.405 | 0.938 | 0.387 | 2.121 | 1.753 |
| Inst.04 | 0.556 | 1.232 | 0.792 | 0.511 | 0.854 | 0.465 | - | - |
| Inst.05 | 0.941 | 0.753 | 0.967 | 0.551 | 0.964 | 0.491 | - | - |

Inst.05 is the most complex problem in the benchmark instances. For this example, a detailed scheduling solution by A2C with soft safety component is found and depicted in Fig. 5.

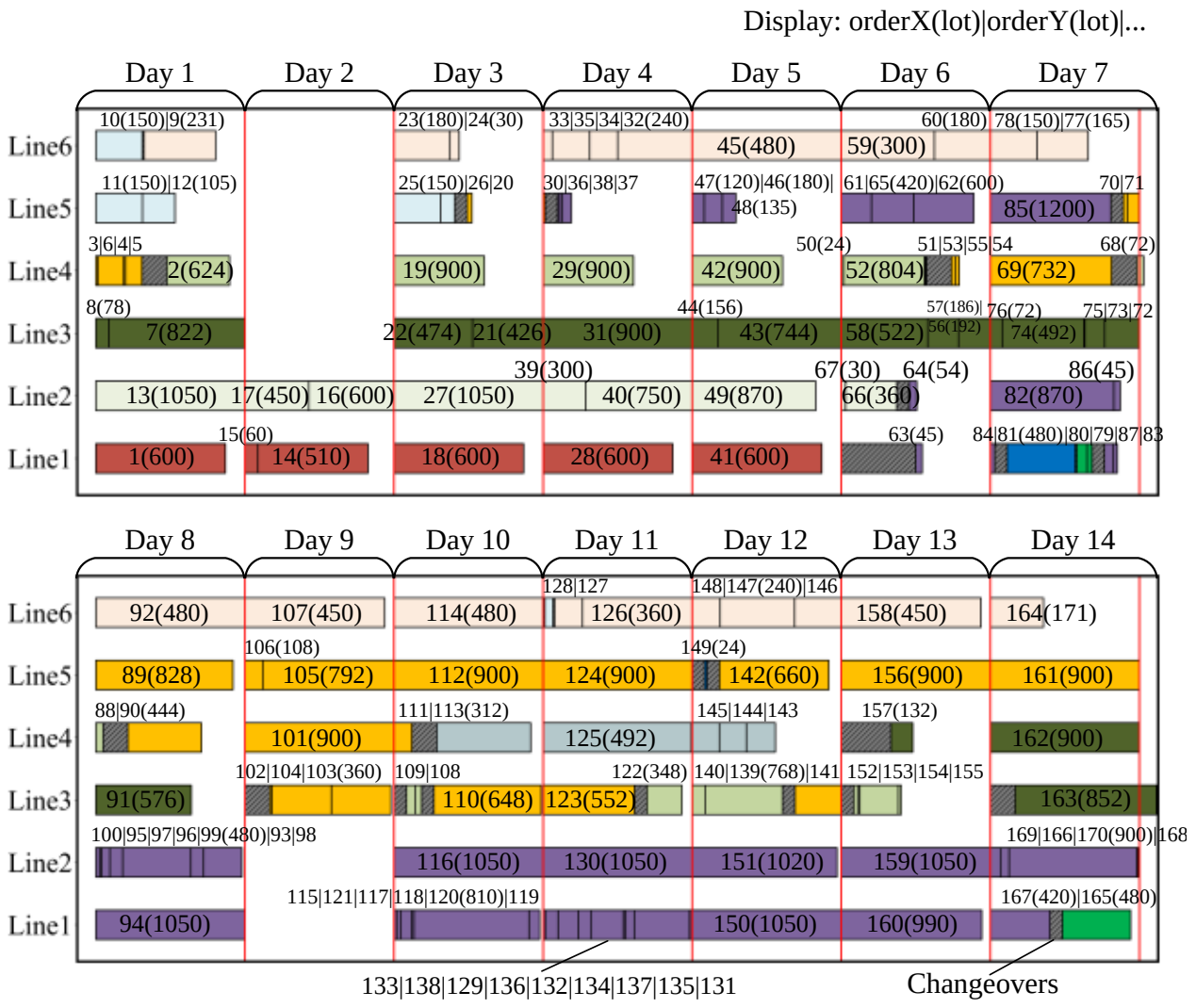

Fig. 5. Scheduling Gantt chart of Inst.05 using A2C with shielding.

The above experiment reveals the CPLEX solver exhibits the lowest efficiency and optimality on Inst.01-Inst.03, and proves incapable of finding a feasible solution on Inst.04 and Inst.05. It can be observed that A2C model provides slightly improved results in addressing the instances with middle to large problem size. Additionally, the integration of the soft safety component in A2C model substantially reduces the total tardiness while slightly losing performance in maintaining low changeover timespan. On the metrics of DCL and DTL, both A2C model and A2C with the soft safety component demonstrate a comparable dispatching ability when compared to INGA in resolving given instances.

## V. CONCLUSIONS

This paper aims to resolve the job scheduling problems in the distributed flexible assembly lines, for which an MILP model is formulated to optimize the changeover time while reducing total tardiness. Beyond that, an advantage actor-critic dispatching model is developed to improve efficiency, reduce tardiness, promote safety and reliability with redesigned representation and encoding strategy for environment state, action space, and reward. On this basis, a Monte Carlo tree search adapted soft shielding component is built to address the unsafe behaviors that may lead to tardiness after a long-sequence dispatching operations. In the experiment, the proposed method as well as its soft safety component is verified in solving benchmark instance and shows reliable applicability.

## Acknowledgment

This work is financially supported by National Key R&D Program of China (Grant 2022YFB4702400).